\documentclass{article}

\usepackage{microtype}
\usepackage{graphicx}
\usepackage{subfigure}
\usepackage{booktabs} 

\usepackage{hyperref}
\usepackage{float}

\usepackage[accepted]{icml2024}

\usepackage{amsmath}
\usepackage{amssymb}
\usepackage{mathtools}
\usepackage{amsthm}

\usepackage[capitalize,noabbrev]{cleveref}

\usepackage{xfrac}

\theoremstyle{plain}

\theoremstyle{definition}

\theoremstyle{remark}

\usepackage[textsize=tiny]{todonotes}

\DeclareMathOperator*{\argmin}{\arg\!\min}

\renewcommand{\vec}[1]{\mathbf{#1}}
\newcommand{\mat}[1]{\mathbf{#1}}
\newcommand{\dif}[0]{\mathrm{d}} 

\newcommand{\KL}{\mathrm{KL}}

\newcommand{\x}{\vec{x}}
\newcommand{\z}{\vec{z}}
\newcommand{\J}{\mat{J}}
\newcommand{\G}{\mat{G}}

\def\T{^{\intercal}} 
\def\R{\mathbb{R}}  
\def\E{\mathbb{E}}  
\def\N{\mathcal{N}} 
 
\newcommand{\M}{\mathcal{M}} 
\newcommand{\Z}{\mathcal{Z}} 
\newcommand{\X}{\mathcal{X}} 

\icmltitlerunning{Decoder ensembling for learned latent geometries}

\begin{document}

\twocolumn[
\icmltitle{Decoder ensembling for learned latent geometries}



\icmlsetsymbol{equal}{*}

\begin{icmlauthorlist}
\icmlauthor{Stas Syrota}{dtu}
\icmlauthor{Pablo Moreno-Muñoz}{dtu}
\icmlauthor{Søren Hauberg}{dtu}
\end{icmlauthorlist}

\icmlaffiliation{dtu}{Department of Applied Mathematics and Computer Science, Technical University of Denmark}

\icmlcorrespondingauthor{Stas Syrota}{stasy@dtu.dk}
\icmlcorrespondingauthor{Pablo Moreno-Muñoz}{pabmo@dtu.dk}
\icmlcorrespondingauthor{Søren Hauberg}{sohau@dtu.dk}

\icmlkeywords{Latent geometries, uncertainty quantification, generative models}

\vskip 0.3in
]



\printAffiliationsAndNotice{}  

\begin{abstract}
  Latent space geometry provides a rigorous and empirically valuable framework for interacting with the latent variables of deep generative models. This approach reinterprets Euclidean latent spaces as Riemannian through a pull-back metric, allowing for a standard differential geometric analysis of the latent space. Unfortunately, data manifolds are generally compact and easily disconnected or filled with holes, suggesting a topological mismatch to the Euclidean latent space.  The most established solution to this mismatch is to let uncertainty be a proxy for topology, but in neural network models, this is often realized through crude heuristics that lack principle and generally do not scale to high-dimensional representations. We propose using ensembles of decoders to capture model uncertainty and show how to easily compute geodesics on the associated expected manifold. Empirically, we find this simple and reliable, thereby coming one step closer to easy-to-use latent geometries.
\end{abstract}

\section{Introduction}
\label{sec:intro}

Generative models provide state-of-the-art density estimators for high-dimensional data \citep{lipman2022flow, sohl2015deep, ho2020denoising, rombach2022high}. In the case of deep latent variable models, such as the \emph{variational autoencoder (\textsc{vae})} \citep{kingma:iclr:2014, rezende:icml:2014}, we assume that data is distributed near a low-dimensional manifold in the spirit of the \emph{manifold hypothesis} \citep{bengio2013representation}. Specifically, we assume that data $\x \in \X$ lies near a low-dimensional manifold $\M \subset \X$, which is parametrized through a low-dimensional \emph{latent representation} $\z \in \Z$. Given finite noisy data, we can recover a stochastic estimate of $\M$. 

Formally, the \textsc{vae} is defined through a (usually unit-Gaussian) prior $p(\z)$ over the latent variables and a conditional likelihood $p(\x | \z)$, which is parametrized by the output of a neural network, $f_{\theta}: \Z \rightarrow \X$, known as the \emph{decoder}. These then define the data density
\begin{align}
  p(\x) = \int p(\x | \z) p(\z) \dif\z.
  \label{eq:vae_likelihood}
\end{align}
Here the latent space $\Z$ is generally Euclidean $\R^d$ with a significantly lower dimension than the observation space $\X$.

We focus on the latent space $\Z$, which generally lacks physical units even when the data may possess such. Following \citet{arvanitidis2021latent}, we consider infinitesimal latent distances measured along the data manifold in observation space. If we let $\z$ denote some latent variable and let $\Delta\z_1$ and $\Delta\z_2$ be infinitesimals, then we can compute the squared distance using Taylor's Theorem,
\begin{align}
  \| f(\z + \Delta\z_1) &- f(\z + \Delta\z_2) \|^2 \\
    &= (\Delta\z_1 - \Delta\z_2)\T \left(\J_{\z}\T \J_{\z}\right) (\Delta\z_1 - \Delta\z_2), \notag
\end{align}
where $\J_{\z} = \frac{\partial f}{\partial \z}\big|_{\z = \z}$ is the Jacobian of the {decoder} $f$.
This implies that the natural distance function in $\Z$ changes locally through the Riemannian metric $\G_\z = \J_{\z}\T \J_{\z}$, which gives the latent space a rich geometric structure.
  
The geometry of the manifold has been shown to carry great value when systematically interacting with the latent representations, as it provides meaningful distances that are independent of how the latent space is parametrized \citep{tosi:uai:2014, arvanitidis2021latent, hauberg:only:2018}. For example, this geometry has allowed \textsc{vae}s to discover latent evolutionary signals in proteins \citep{detlefsens:proteins:2020}, provide efficient robot controls \citep{scannellTrajectory2021, chen2018active, hadi:rss:2021}, improve latent clustering abilities \citep{yang:arxiv:2018, arvanitidis2021latent} and more.

The fundamental issue with these geometric approaches is that by assuming the latent space to have an Euclidean topology, we impose the same topology on the manifold $\M$ in observation space. In practice, we have little \emph{a priori} information about the topology of the true manifold and must rely on the observed data to estimate a reasonable topology. As data is finite, we should expect such an estimate to be compact, and empirically it is often observed that manifolds arising from real-world data are disconnected and often have holes. All of which mismatches the Euclidean latent topology. 

\citet{hauberg:only:2018} argues that the uncertainty of the decoder offers a \emph{topological hint}, i.e.\@ when model uncertainty is high we are most likely outside the support of the driving manifold. When the decoder follows a \emph{Gaussian process (GP)}, there is a well-established notion of model uncertainty, and its impact on the latent geometry is reasonably well-understood \citep{tosi:uai:2014, pouplin:finsler:2023}. However, when the decoder is a neural network (the ever-present case), a set of heuristics is commonly applied to mimic the behavior of the GP models \citep{arvanitidis2021latent, arvanitidis:aistats:2019, arvanitidis:aistats:2021, arvanitidis:aistats:2022, detlefsen:2019:reliable, detlefsens:proteins:2020,hadi:rss:2021}. Besides lacking principle, these heuristics also tend to break down when the latent dimension exceeds a handful.

\textbf{In this paper}, we propose to use an ensemble \citep{lakshminarayanan2017simple, hansen1990neural} of decoders in the \textsc{vae} to capture model uncertainty and provide simple training techniques for the associated model. We then show how to easily incorporate the ensemble into the computation of geodesics on the modeled stochastic manifold. The result is a simple, yet reliable, approach for leveraging uncertainty in learned geometric representations. 

\section{Background and related work}
  \subsection{Variational autoencoders}
    We briefly review the variational autoencoder (\textsc{vae}) as our empirical results are realized with this generative model. Many of our findings, however, extend beyond this model.

    The \textsc{vae} \citep{kingma:iclr:2014, rezende:icml:2014} is a deep latent variable model that generalizes \emph{probabilistic principal component analysis} \citep{tipping1999probabilistic}. Commonly the latent variable is assumed \emph{a priori} to follow a unit-Gaussian, $p(\z) = \N(\z | \vec{0}, \mat{I})$, though more elaborate priors have been studied \citep{tomczak2018vae, kalatzis:icml:2020, rombach2022high}. The conditional likelihood $p(\x | \z)$ is then parametrized by a neural network $f_{\theta}(\z)$ known as the \emph{decoder}. For example, for a Gaussian \textsc{vae}, we let $f_{\theta}(\z) = (\mu_{\theta}(\z); \sigma_{\theta}(\z))$ and
    \begin{align}
      p(\x | \z) &= \N(\x | \mu(\z), \sigma^2(\z)\mat{I}),
    \end{align}
    where we omitted the $\theta$ subscript for brevity. The data likelihood \eqref{eq:vae_likelihood} arise by the marginalization of $\z$, but, alas, the associated integral is generally intractable and we resort to a lower bound, known as the \emph{ELBO}, \citep{kingma:iclr:2014, rezende:icml:2014}
    \begin{align}
      \mathcal{L}_{\theta, \psi} =
      \E_{q_{\psi}(\z | \x)} \left[ \log p_{\theta}(\x | \z) \right] - \KL(q_{\psi}(\z | \x) \| p(\z)),
      \label{eq:vae_elbo}
    \end{align}
    where $q_{\psi}(\z | \x) = \N(\z | \mu_{\psi}(\x), \sigma^2_{\psi}(\x))$ is a variational approximation to the latent posterior $p(\z | \x)$. Details can be found in the original papers.

    Despite mentioning priors and posteriors, the \textsc{vae} is inherently non-Bayesian as it relies on maximum likelihood to arrive at a point estimate of the decoder parameters $\theta$. \citet{daxberger2019bayesian} gives the model a Bayesian treatment and relies on stochastic gradient Markov chain Monte Carlo for inference.

    We focus on the common case where the latent space is assumed to have an Euclidean structure, i.e.\@ $\Z = \R^d$. This is, however, not a strict requirement and other latent structures have been investigated \citep{davidson2018hyperspherical, mathieu2019continuous}.

\subsection{Latent representation geometries}
  The latent variables of the \textsc{vae} are enticing as they provide low-dimensional `distillations' of high-dimensional data. This form of representation learning \citep{bengio2013representation} can give us a \emph{glimpse} into the model's inner workings, but also potentially in the mechanisms of the true physical system that generated the data.
  
  Unfortunately, the latent space can be almost arbitrarily deformed without changing the associated model density \citep{hauberg:only:2018}. To see this, consider a smooth invertible function $h: \Z \leftarrow \Z$, such that its inverse is also smooth (i.e.\@ a \emph{diffeomorphism}). If the Jacobian of $h$ further has unit determinant, we see that the latent representations $\hat{\z} = h(\z)$ yields an unchanged density when combined with the decoder $\hat{f} = f \circ h^{-1}$. This implies that whichever latent representations we may recover from optimizing Eq.~\ref{eq:vae_elbo}, are not unique. This lack of uniqueness hinders any form of interpretability of the latent representations, such that the aforementioned `glimpse' becomes difficult to trust. \citet{hauberg:dggm:2023} discuss this issue at greater length, while \citet{detlefsens:proteins:2020} show the empirical significance of the problem in a model of proteins.

  Fortunately, \emph{differential geometry} provides an elegant solution \citep{arvanitidis2021latent, shao2018riemannian}. The basic idea is to define distances in the latent space by measuring infinitesimally along the spanned manifold in observation space. Specifically, consider a latent curve $\gamma: [0, 1] \rightarrow \Z$ and its decoded counterpart $f \circ \gamma: [0, 1] \rightarrow \X$. We may then define the length of $\gamma$ by integrating $f \circ \gamma$, i.e.
  \begin{align}
    \mathrm{Length}[\gamma] &= \int_0^1 \left\| \frac{\dif}{\dif t} f(\gamma_t) \right\| \dif t,
  \end{align}
  where $\gamma_t = \gamma(t)$. Applying the chain rule quickly reveals that the integrand can be written as
  \begin{align}
    \left\| \frac{\dif}{\dif t} f(\gamma_t) \right\|
      &= \sqrt{ \dot{\gamma}_t\T \J_{\gamma_t}\T \J_{\gamma_t} \dot{\gamma}_t },
  \end{align}
  where $\dot{\gamma}_t = \sfrac{\dif \gamma}{\dif t} |_{t=t}$ denotes the curve derivative. The matrix $\J_{\gamma_t}\T \J_{\gamma_t}$, thus, defines a local inner product, which is known as a \emph{Riemannian metric}. From this notion of curve length, we can define the associated distance that measures the length of the shortest path, also known as the \emph{geodesic},
  \begin{align}
    \mathrm{dist}(\z_0, \z_1)
      &= \mathrm{Length}[\gamma^*], \qquad\text{where} \\
    \begin{split}
    \gamma^*
      &= \argmin_{\gamma} \mathrm{Length}[\gamma] \\
      & \text{s.t.} \quad \gamma_0 = \z_0 \text{ and } \gamma_1 = \z_1.
    \end{split}
  \end{align}
  This distance measure does not change if we reparametrize the latent space by some diffeomorphism $h: \Z \rightarrow \Z$. This construction can be expanded upon to allow for reparametrization invariant measurements of volumes, angles, and more (see \citet{hauberg:dggm:2023} for details).

\subsection{Topology estimation and the role of uncertainty}
  Training data is inherently finite, suggesting that we should only expect to be able to learn a compact manifold. Further, it is not unreasonable to expect that the underlying manifold near which the data are distributed can have holes. These considerations lead to a mismatch between the topology of the manifold we seek to estimate and the Euclidean topology of the latent space.

  In rare cases, we may have prior topological information about the underlying manifold and we can adapt the latent space accordingly \citep{davidson2018hyperspherical, mathieu2019continuous}. Generally, we, however, must estimate the underlying manifold's topology if we are to reliably estimate its geometry.\looseness=-1

  One approach to topology estimation is to cover the manifold using multiple charts and learn diffeomorphisms that connect these \citep{kalatzis:arxiv:2021, schonsheck2019chart}. This, however, notably complicates model estimation, and the approach is rarely followed in practice.

  \begin{figure}
      \centering
      \includegraphics[width=0.7\linewidth]{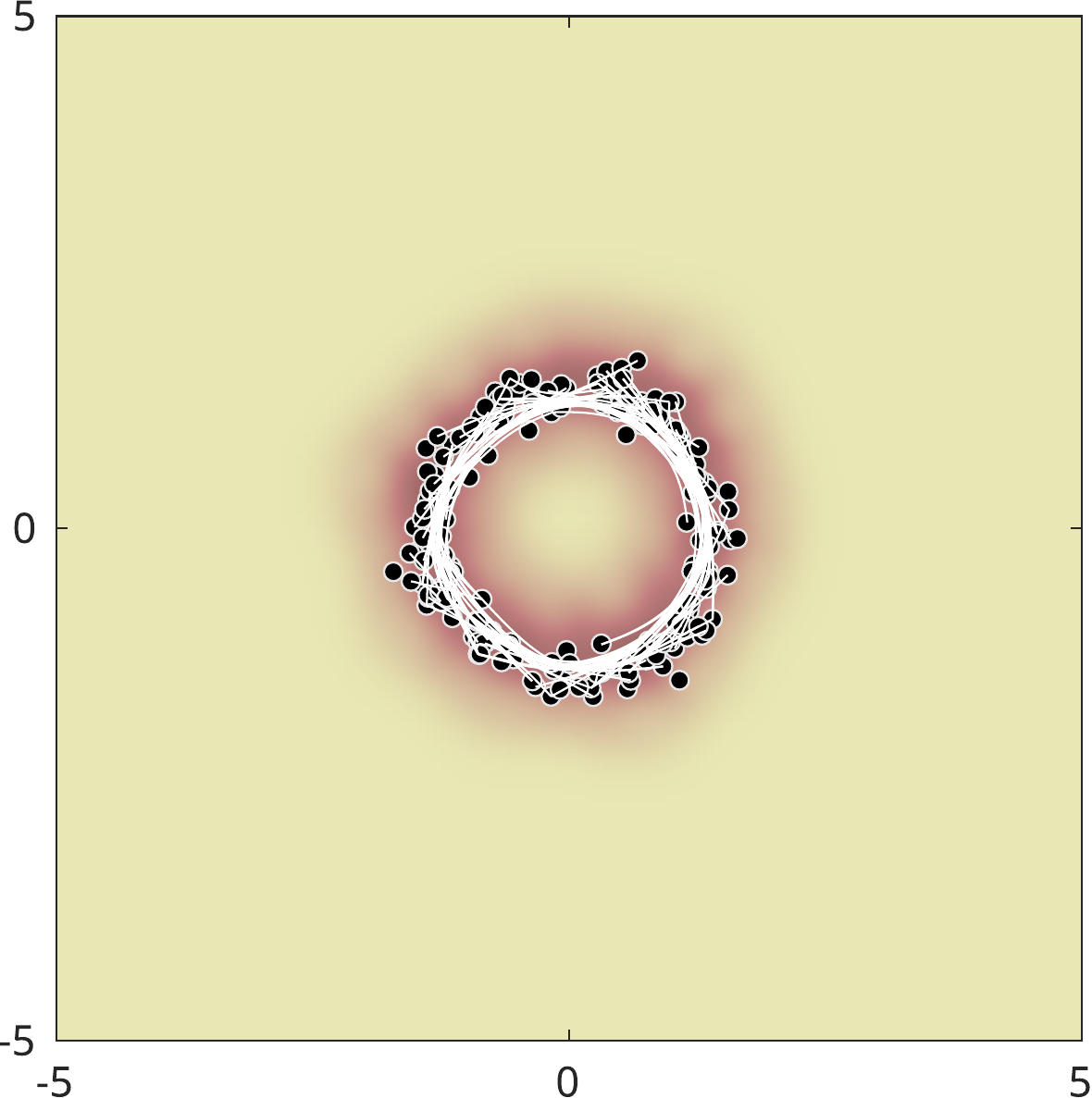}
      \caption{Shortest paths (geodesics) under the expected metric of a decoder following a Gaussian process. The topological hint of uncertainty is, thus, propagated to the metric. \emph{Figure is courtesy of \citet{hauberg:only:2018}}.}
      \label{fig:circle}
  \end{figure}
  \citet{hauberg:only:2018} argues that model uncertainty offers a topological hint. The intuition is that the decoder should have high uncertainty in regions of the latent space with little support from training data (i.e.\@ \emph{outside the manifold}). One approach to incorporating model uncertainty into the geometry is to consider the \emph{expected Riemannian metric} \citep{tosi:uai:2014},
  \begin{align}
    \E[\mat{G}_{\z}]
      &= \E[\J_{\z}\T \J_{\z}]
       = \E[\J_{\z}]\T \E[\J_{\z}] + \mathrm{cov}(\J_{\z})
  \end{align}
  such that distances are larger in regions of high uncertainty \citep{hauberg:only:2018}. The property ensures that geodesics stay close to the training data (Fig.~\ref{fig:circle}).
  The expected metric has been analyzed in great detail when the decoder follows a posterior Gaussian process \citep{pouplin:finsler:2023}.

  To the best of our knowledge, the expected metric has only been explored for decoders following Gaussian processes and not for neural network decoders. To shape the metric to take large values outside the data support, \citet{arvanitidis2021latent} suggests taking the variance of the conditional likelihood $p(\x | \z)$ into account. Specifically, for a Gaussian conditional likelihood, \citeauthor{arvanitidis2021latent} suggests the metric
  \begin{align}
    \mat{G} &= \J_{\mu}\T\J_{\mu} + \J_{\sigma}\T\J_{\sigma}.
  \end{align}
  Assuming $\sigma^2(\z)$ grows with the distance to training data, then this metric will give rise to geodesics that approach the data. Unfortunately, the neural network $\sigma: \Z \rightarrow \X$ does not exhibit such growth on its own, and \citet{arvanitidis2021latent} heuristically proposed to model $\sigma^{-2}$ with a \emph{radial basis function neural network} \citep{que2016back}, which provides such growth. Variants of this heuristic are commonly applied when using learned latent geometries \citep{arvanitidis:aistats:2022, detlefsens:proteins:2020, detlefsen:2019:reliable, hadi:rss:2021}.

\subsection{Computing geodesics}
  There are several ways to compute the geodesic that connects two points. The classic approach amounts to solving the geodesic differential equation as a two-point boundary value problem \citep{hauberg:nips:2012, arvanitidis:aistats:2019, miller2006geodesic}. This works well for low-curvature manifolds, such as spheres and tori, but is generally unstable on learned manifolds. On low-dimensional manifolds, we can alternatively discretize the manifold into a graph and apply classic algorithms for computing shortest paths on graphs \citep{hadi:rss:2021}. The size of such a graph, however, grows exponentially with the manifold dimension, and the approach is impractical beyond three dimensions.

  A more practical approach is to note that minimizers of \emph{curve length} coincides with those of \emph{curve energy} \citep{carmo1992riemannian},
  \begin{align}
    \mathcal{E}[\gamma] &= \int_0^1 \left\| \frac{\dif}{\dif t} f(\gamma_t) \right\|^2 \dif t,
    \label{eq:energy}
  \end{align}
  which follows from the Cauchy-Schwarz inequality. Minimizing curve energy has the benefit of yielding solution curves with constant speed \citep{carmo1992riemannian}. This energy can easily be discretized as
  \begin{align}
    \mathcal{E}[\gamma] &\approx \sum_{t=0}^{T-1} \left\| f(\gamma(\sfrac{t+1}{T})) - f(\gamma(\sfrac{t}{T})) \right\|^2 \dif t.
    \label{eq:disc_energy}
  \end{align}

  A simple algorithm for computing geodesics then amounts to parametrizing the curve $\gamma$ and minimizing the discretized energy \eqref{eq:disc_energy} with respect to the curve parameters. \citet{shao2018riemannian} propose parametrizing the curve as a discrete set of points, \citet{yang:arxiv:2018} use a second-order polynomial, while \citet{software:stochman} use cubic splines. In our implementation, we opt for the latter.

\section{Ensemble of decoders}
  To capture the model uncertainty of a \textsc{vae}, we need to access the posterior distribution over the model parameters $\theta$. Since the encoder is not part of the model, but rather an amortization mechanism for the variational inference, we are only interested in the posterior of the decoder weights, $p(\theta | \mathcal{D})$, where $\mathcal{D}$ denotes the training data.

  In practice, current Bayesian deep learning techniques often struggle to approximate the posterior over the weights. We, therefore, propose to approximate posterior samples with a \emph{deep ensemble} \citep{lakshminarayanan2017simple}, which can, heuristically, be seen as a Bayesian approximation \citep{gustafsson2020evaluating}.

  This can trivially be implemented by instantiating $S$ randomly initialized decoders $\{ f_{\theta_s} \}_{s=1}^S$. For each mini-batch of data, we randomly sample a decoder $f_{\theta_s}$ and take a gradient step to optimize the ELBO $\mathcal{L}_{\theta_s, \psi}$. At convergence, we have access to one encoder and $S$ decoders.

\begin{figure}[h!]
      \centering
      \includegraphics[width=\columnwidth]{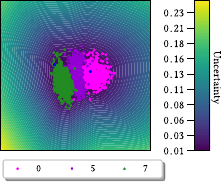}
      \caption{Using an ensemble of decoders ensures that regions of the latent space with limited data support have high uncertainty.}
      \label{fig:ensemble_uq}
  \end{figure}

  Figure~\ref{fig:ensemble_uq} shows an example of the uncertainty of an ensemble of decoders. For ease of visualization, we consider a \textsc{vae} with a two-dimensional latent space trained on three classes of \textsc{mnist}. We show the uncertainty in the latent space, which we have calculated as the mean over $n$ pixel standard deviations
  \begin{align}
      \text{uncertainty}(\z^{\prime})=\frac{1}{n}\sum_{i=1}^n \sigma_i(\z^{\prime})
  \end{align}
  with 

\begin{align*}
      \sigma(\z^{\prime}) &= \sqrt{\frac{1}{S} \sum_{j=1}^{S} (f_{\theta_j}(\z^{\prime})- \mu(\z^{\prime}))^2}\\
      \mu(\z^{\prime})&= \frac{1}{S}\sum_{j=1}^{S}f_{\theta_j}(\z^{\prime})
  \end{align*}

  We see that uncertainty generally grows with the distance to the latent representations as one would naturally expect. This gives hope that this uncertainty can be used to shape the Riemannian metric to better respect topology.

\section{Ensemble geodesics}
  To obtain a practical latent geometry from the decoder ensemble, we think of this as samples from an approximate posterior $f_{\theta} \sim q(\theta)$. We may, thus, construct an \emph{expected metric} as $\mat{G} = \E_{q(\theta)}[\J_{f_{\theta}}\T \J_{f_{\theta}}]$. Under this metric, we see that the energy of a curve $\gamma$ becomes
  \begin{align}
    \mathcal{E}[\gamma]
      &= \int_0^1 \E_{q(\theta)}\left[ \dot{\gamma}_t\T \J_{f_{\theta}}\T \J_{f_{\theta}} \dot{\gamma}_t \right] \dif t,
  \end{align}
  and following the discretization of Eq.~\ref{eq:disc_energy} we get
  \begin{align*}
    \mathcal{E}[\gamma] &\approx \sum_{t=0}^{T-1} \E_{q(\theta)} \left[ \left\| f_{\theta}(\gamma(\sfrac{t+1}{T})) - f_{\theta}(\gamma(\sfrac{t}{T})) \right\|^2 \right].
  \end{align*}

  \begin{figure*}[h]
    \centering
    \includegraphics[width=\textwidth]{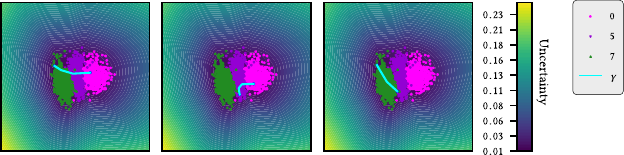}
    \includegraphics[width=\textwidth]{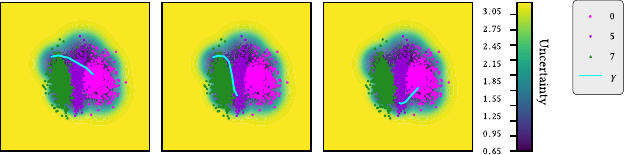}
    \caption{\textbf{Upper row:} Three examples of the latent space for ensembles of \textsc{vae} decoders on a reduced version of \textsc{mnist} data with \emph{three} classes. Blue curves indicate the geodesic interpolants between two random latent coordinates. 
    \textbf{Lower row:} Three examples of the latent space for a \textsc{vae} with \textsc{rbf}-generated uncertainties on \textsc{mnist} data with \emph{three} classes.}
    \label{fig:rbf}
\end{figure*}

  Empirically, we have found that geodesics that minimize this discretized energy do not follow the data as closely as one could hope for. We hypothesize that the decoder ensemble \emph{underestimate} model uncertainty since all ensemble members are trained always on the same data points. Similar issues have been previously observed with other classical ensemble-based models, i.e. \emph{bootstrap} methods \citep{efron1983leisurely}. To counter this, we will modify the energy to amplify the impact of model uncertainty by disregarding correlations.

  Particularly, the discretized energy sums expected squared distances $\E[\|f(\z_2) - f(\z_1)\|^2]$. To analyze this, we introduce the short-hand notation $\x_i =f_{\theta}(\z_i)$ and write the moments of a pair $(\x_1, \x_2)$ as
  \begin{align}
    \E_{q(\theta)}\left[ \begin{pmatrix}
      \x_1 \\ \x_2
    \end{pmatrix} \right]
    &= 
    \begin{pmatrix}
      \mu_1 \\ \mu_2
    \end{pmatrix},
    \\
    \mathrm{cov}\left[ \begin{pmatrix}
      \x_1 \\ \x_2
    \end{pmatrix} \right]
    &= 
    \begin{pmatrix}
      \Sigma_{11}  & \Sigma_{12} \\
      \Sigma_{12}  & \Sigma_{22}
    \end{pmatrix}.
  \end{align}
  The difference vector $\Delta = \x_2 - \x_1$ will thus have moments
  \begin{align}
    \E[\Delta] &= \E[\x_2] - \E[\x_1], \\
    \mathrm{cov}[\Delta] &= \Sigma_{11} + \Sigma_{22} - 2 \Sigma_{12},
  \end{align}
  and the individual summands of the discretized energy are then of the following form
  \begin{align}
    \E\left[ \| \Delta \|^2 \right]
      &= \| \E[\Delta] \|^2 + \mathrm{tr}[\Sigma_{11} + \Sigma_{22}] - 2 \mathrm{tr}[\Sigma_{12}].
      \label{eq:crosscov}
  \end{align}
  We explicate these expressions to emphasize that cross-covariances between points along $\gamma$ decrease the curve energy. Neural network ensembles are known to provide better performance under \emph{de-correlated} predictions, which is \emph{de facto} a way to promote higher degrees of ensemble diversity \citep{lakshminarayanan2017simple,lee2016stochastic}. 
  
\begin{figure}[H]
  \centering
  \includegraphics[width=0.85\columnwidth]{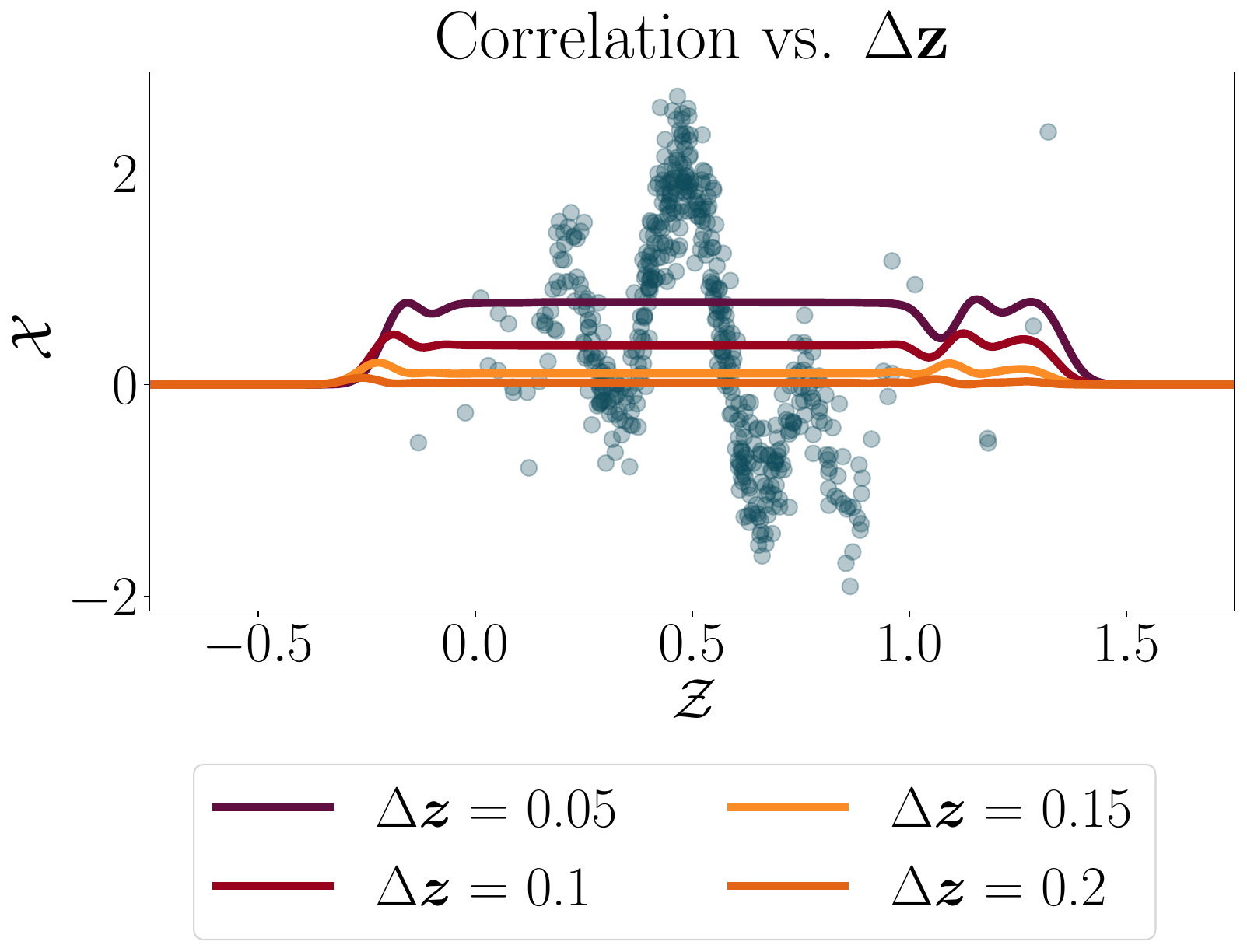}
  \caption{The \emph{correction} term of posterior covariances in a GP tends to be zero as $\Delta\mathbf{z}\gg 0$, even in areas of $\mathcal{Z}$ where is a high-density of training data.}
  \label{fig:correlations}
\end{figure}

  Additionally, the correlation terms in posterior cross-covariances in other probabilistic models like \emph{Gaussian processes} \citep{williams2006gaussian}, also collapse to zero values as the size of difference vector $\Delta$ augments (see Figure~\ref{fig:correlations}). This primarily indicates that the discretized energy can be also negatively affected by spurious cross-covariance terms whenever the difference is not sufficiently small given the high flexibility of the ensemble neural networks.

  In practice, all of this suggests that cross-covariance terms like $\Sigma_{12}$ in Eq.~\ref{eq:crosscov} are not beneficial for the minimization of the discretized energy with ensembles and we drop them, i.e.
  \begin{align}
    \E\left[ \| \Delta \|^2 \right]
      &\approx \| \E[\Delta] \|^2 + \mathrm{tr}[\Sigma_{11} + \Sigma_{22}].
  \end{align}
  This can be practically implemented by evaluating the energy directly as
  \begin{align*}
    \mathcal{E}[\gamma] &\!\approx\! \sum_{t=0}^{T-1} \E_{\theta, \theta' \sim q(\theta)q(\theta)} \left[ \left\| f_{\theta}(\gamma(\sfrac{t+1}{T})) \!-\! f_{\theta'}(\gamma(\sfrac{t}{T})) \right\|^2 \right]
  \end{align*}
  When minimizing this energy, we use a simple one-sample Monte Carlo estimate.




\section{Experiments}
We compare our method to the current state-of-the-art approach to model uncertainty when learning latent geometries \citet{arvanitidis2021latent}, which relies on \textsc{rbf} networks to model data uncertainty. In particular, we show that geodesic distances stemming from our ensemble of decoders method are more stable under retraining when compared to distances learned using the \textsc{rbf} neural network. The implementation of our method and code producing the expreimental results is available at \href{https://github.com/mustass/ensertainty}{https://github.com/mustass/ensertainty}.

For both approaches, we choose a \textsc{vae} architecture with dense layers whereas for the \textsc{rbf} neural network part we use a mixture of 10 Gaussians in the latent space. We train the models on the \textsc{mnist} and \textsc{fmnist} datasets with two-dimensional latent spaces. We further extend the \textsc{mnist} analysis to 50-dimensional latents.

 We retrain the \textsc{vae}s using 30 different seeds on both datasets. We subsequently calculate the geodesic distances between the latent representations of 100 pairs of randomly chosen data points from the test set. These points are fixed across all trials. The outcome is 30 measurements per point pair for both methods. This allows us to calculate the coefficient of variation (CV) for each method to compare their variability and, thus, robustness,
\begin{equation}
    \text{CV} = \frac{\sigma}{\mu},
    \label{eq:cv}
\end{equation}
where $\mu$ and $\sigma$ are the mean and standard deviations of the distances calculated for the same point pair by 30 different estimations of a model. Note that CV is a unitless measure of relative variability, where a lower value indicates less variability and, thus, allows us to compare the variability of values on different scales.

\begin{figure*}[h]
    \centering
    \includegraphics[width=\columnwidth]{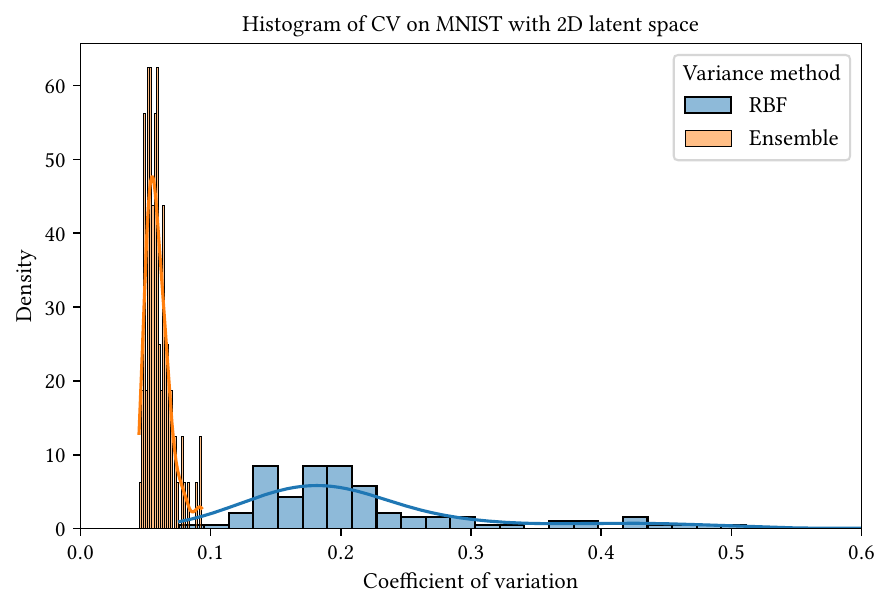}
    \includegraphics[width=\columnwidth]{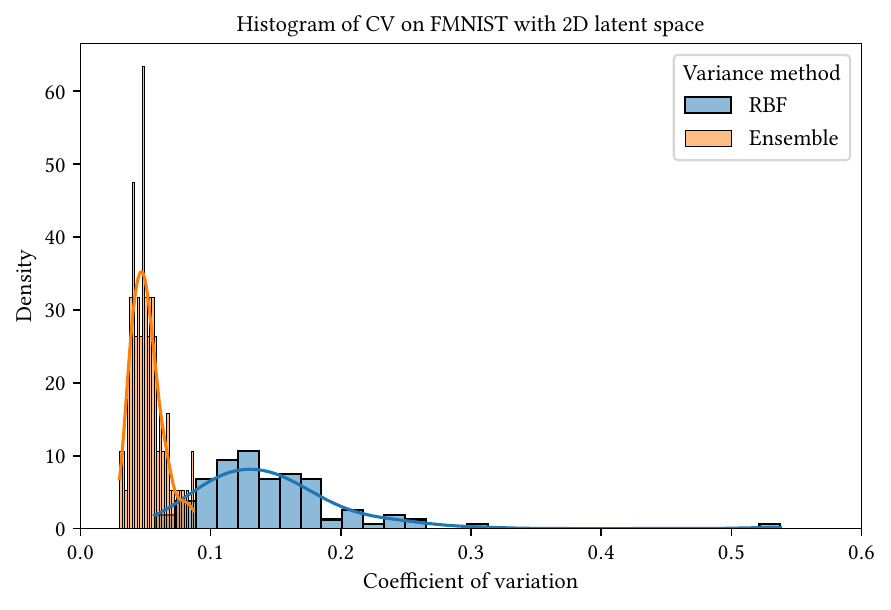}
    \caption{Histogram of coefficients of variation for \textsc{mnist} and \textsc{fmnist} data with $d=2$ in the latent space $\mathcal{Z}$.}
    \label{fig:histogram-mnist}
\end{figure*}

Table~\ref{tab:ttests} shows results for the one-sided paired Student's $t$-test with the null hypothesis of ensemble geodesics having a lower coefficient of variation than by using the \textsc{rbf}-based model. The results show that the ensemble of decoders is consistently more reliable than the \textsc{rbf}-based approach, which is currently the most popular approach. Figure~\ref{fig:histogram-mnist} visualizes the findings using a histogram of coefficients of variation for different point pairs.

\begin{table*}[ht]
\centering
    \begin{tabular}{rrlllrr}
    \toprule
    Dataset & Number of classes & $d$ & Null-hypothesis & Alternative & $t$-statistic & $p$-value \\
    \midrule
    \textsc{mnist-3} & 3  & 2 & Ensemble geodesics have lower CV & greater & -16.834 & 1.000 \\
    \textsc{mnist} & 10 & 2 & Ensemble geodesics have lower CV  & greater & -16.290 & 1.000 \\
    \textsc{fmnist} & 10 & 2 &   Ensemble geodesics have lower CV  & greater & -15.339  & 1.000 \\
    \textsc{mnist} & 10 & 50 & Ensemble geodesics have lower CV & greater & -6.472 & 0.999 \\
    \bottomrule
    \end{tabular}
    \caption{Statistical metrics ($p$-value and $t$-statistic) for \textsc{mnist} and \textsc{fmnist} with ensembles and \textsc{rbf} uncertainties.}
    \label{tab:ttests}
\end{table*}

\section{Conclusion}
  Learned latent geometries crucially rely on uncertainty estimation in order to shape the metric according to the underlying manifold's topology \citep{hauberg:only:2018}. In Gaussian process models \citep{tosi:uai:2014, pouplin:finsler:2023} this construction naturally comes in place, but models based on neural networks have required a series of heuristics to behave desirable \citep{arvanitidis2021latent}. Unfortunately, these heuristics work poorly beyond a few latent dimensions.

  We have proposed to use neural network ensembles to capture model uncertainty. We have shown that this leads to empirical improvements compared to current heuristics. In practice, training ensembles of decoders requires only small code modifications and our proposed approach is generally easy to implement.

\section*{Acknowledgments}
This work was supported by a research grant (42062) from VILLUM FONDEN. This project received
funding from the European Research Council (ERC) under the European Union’s Horizon 2020
research and innovation programme (grant agreement 757360). The work was partly funded by the Novo Nordisk Foundation through the Center for Basic Machine Learning Research in Life Science
(NNF20OC0062606).
\newpage

\bibliography{main.bib}
\bibliographystyle{icml2024}




\end{document}